\documentclass[10pt,twocolumn,letterpaper]{article}
\usepackage[utf8]{inputenc}
\usepackage{siunitx}

\usepackage{cvpr}
\usepackage{times}
\usepackage{epsfig}
\usepackage{graphicx}
\usepackage{amsmath}
\usepackage{mathptmx}
\usepackage{amssymb}
\usepackage{url}
\usepackage{subcaption}
\usepackage{fancyref}
\usepackage{color}
\usepackage{booktabs} 
\usepackage{textcomp}
\usepackage{stackengine}

\usepackage[pagebackref=true,breaklinks=true,letterpaper=true,colorlinks,bookmarks=false]{hyperref}

 \cvprfinalcopy 

\def\cvprPaperID{3882} 

\ifcvprfinal\fi
\begin{document}

\setlength{\abovedisplayskip}{3pt}
\setlength{\belowdisplayskip}{3pt}

\title{Hands: An Important Cue in Egocentric Images}

\title{An analysis of hand segmentation in first- and third-person vision}

\title{Analysis of Hand Segmentation in the Wild}

\author{Aisha Urooj Khan and Ali Borji \\
Center for Research in Computer Vision, University of Central Florida\\
{\tt\small \{aishaurooj,aliborji\}@gmail.com}
}

\maketitle

\begin{abstract}
\vspace*{-5pt}
   	A large number of works in egocentric vision have concentrated on action and object recognition. Detection and segmentation of hands in first-person videos, however, has less been explored. For many applications in this domain, it is necessary to accurately segment not only hands of the camera wearer but also the hands of others with whom he is interacting. Here, we take an in-depth look at the hand segmentation problem. In the quest for robust hand segmentation methods, we evaluated the performance of the state of the art semantic segmentation methods, off the shelf and fine-tuned, on existing datasets. We fine-tune RefineNet, a leading semantic segmentation method, for hand segmentation and find that it does much better than the best contenders. Existing hand segmentation datasets are collected in the laboratory settings. To overcome this limitation, we contribute by collecting two new datasets: a) EgoYouTubeHands including egocentric videos containing hands in the wild, and b) HandOverFace to analyze the performance of our models in presence of similar appearance occlusions. 
We further explore whether conditional random fields can help refine generated hand segmentations. To demonstrate the benefit of accurate hand maps, we train a CNN for hand-based activity recognition and achieve higher accuracy when a CNN was trained using hand maps produced by the fine-tuned RefineNet. Finally, we annotate a subset of the EgoHands dataset for fine-grained action recognition and show that an accuracy of 58.6\% can be achieved by just looking at a single hand pose which is much better than the chance level (12.5\%).

\end{abstract}
\vspace*{-10pt}
\section{Introduction}
\vspace*{-5pt}
Growing usage of wearable devices such as Google Glass, GoPro, and Narrative Clip has made egocentric research in computer vision a rapidly growing area. These cameras generate huge volumes of data which makes automatic analysis of their recorded content (e.g., for browsing, searching, and visualizing) a need, for applications such as summarizing videos, describing events in life-logging photo data, and recognizing activities of daily living. Most of the work in egocentric vision deals with understanding camera wearer's activities and behavior. In this work, instead, we focus on a very crucial entity in egocentric videos: \emph{hands}. Hands are ubiquitous in our daily life. We see them more than any other object in our life time. Their pose and configuration tell a lot about what we plan to do or what we pay attention to. Due to these, hand detection, segmentation and tracking are fundamental problems in egocentric vision with a myriad of applications in robotics, human-machine interaction, computer vision, augmented reality, etc. Extracting hand regions in egocentric videos is a critical step for understanding fine motor skills such as hand-object manipulation and hand-eye coordination.

We address the task of egocentric pixel-level hand detection and segmentation in realistic daily settings. A large number of works have addressed this problem in third-person or surveillance videos. Relatively less effort, however, has been devoted to this problem in first-person videos. Although few works exist (e.g., \cite{Betancourt2016}, \cite{li2013pixel} and \cite{egohands2015iccv}), the last analysis of hand segmentation in egocentric videos dates back to pre deep learning era~\cite{li2013pixel}. Here, we plan to renovate this topic by conducting an exhaustive analysis of hand segmentation in egocentric videos using state of the art semantic segmentation methods.  

In contrast to third-person point-of-view videos -- e.g., from a mounted surveillance camera or a laptop camera -- egocentric videos contain rapid changes in illumination, highly dynamic and unpredictable camera motion, unusual composition and viewpoints, significant motion blur, and complex hand-object manipulations. Further, camera wearer is not captured in egocentric videos, thus other cues that can localize hands might be missing (e.g., person's face or shoulders). Thus, hand segmentation models developed over third-person videos may not be adequate for egocentric hand detection.





We base our analysis on Bambach et al.~\cite{egohands2015iccv}, where they introduced a new hand dataset and a deep learning model for hand detection and segmentation. Their dataset, EgoHands, has pixel-level annotations for hands with two participants in each video interacting with each other \cite{egohands2015iccv}. We chose this dataset for two main reasons: 1) To the best of our knowledge, it is the only egocentric dataset with focus on humans interactions from first-person point-of-view, and 2) It has pixel-level annotations for hands, where hands are considered from fingers till wrist. We also utilize the GTEA dataset which includes cooking activities indoors. Further, we introduce a new large-scale dataset of labeled hands in the wild which includes YouTube videos shot in realistic unconstrained conditions, indoors and outdoors, under a wide variety of scenarios. We call this dataset as EgoYouTubeHands (EYTH) dataset. 
An analysis is also performed to assess how general the segmentation model could be by applying them to hands across datasets(both first-person and third-person images).

We also study the utility of hand regions for activity recognition, hands alone and in conjunction with  manipulated objects.

In summary, the lessons learned from this study are as follows: 
\textbf{First,} we find that fine-tuning the RefineNet model for pixel-level hand detection dramatically improves the state of the art (e.g., 81.4\% mIOU over EgoHands dataset, about 26\% improvement vs. \cite{egohands2015iccv}), \textbf{Second,} we annotate hands at pixel-level over approx. 1300 egocentric video frames taken in unconstrained real world environments and evaluated RefineNet on those images.  
We find that finetuning RefineNet over these images generalizes well across other datasets in terms of mIOU accuracy, \textbf{Third,} we introduce a new HandOverFace dataset which has 300 frames with faces occluded by hands to test the performance of hand segmentation methods, \textbf{Fourth,} we applied the Conditional Random Fields to refine the model segmentation maps and found that it improves the accuracy in some cases, although not always,  \textbf{Fifth,} we annotated a subset of EgoHands dataset (800 frames) for finer hand-level actions like picking, placing, holding, etc. and found that a single hand pose carry much information about activity being performed, and \textbf{Finally,} we trained AlexNet on that subset 
and achieved 58.6\% accuracy (chance equals 12.5\% for 8 most frequent action classes), and 59.2\% accuracy when finely annotated hand maps were used. We find that hand maps along with objects significantly improves activity recognition (77.3\% recognition accuracy).

\section{Related Works} \label{related_work}
\vspace*{-5pt}

\noindent \textbf{First-person hand segmentation.} 
Li and Kitani et al.,~\cite{li2013pixel} classified hand detection approaches (pre deep learning era models) into three categories: (1) local appearance-based detection; e.g., those relying on skin color~\cite{jones2002statistical,kakumanu2007survey,argyros2004real}, (2) global appearance-based detection; e.g., using global hand template~\cite{sudderth2004visual,rehg1994visual,stenger2001model}, and (3) motion-based detection; using 
ego-motion of the camera and assuming hands (foreground) and the background have different motion~\cite{sheikh2009background,hayman2003statistical,fathi2011learning}. They also presented a dataset of over 600 hand images taken under various illumination and backgrounds and highlight the pros and cons of various hand detection methods.

\begin{figure*}[t]

\begin{tabular}{cccc}
\subcaptionbox{EgoHands}{\includegraphics[width=0.23\linewidth]{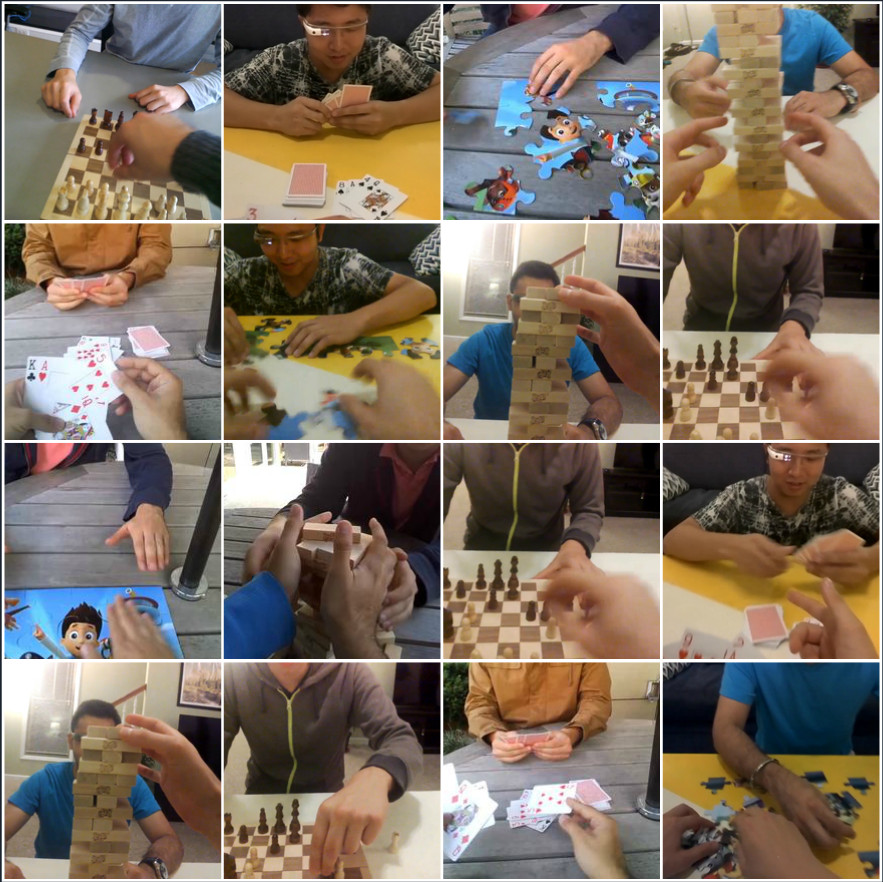}} &
\subcaptionbox{EgoYouTubeHands}{\includegraphics[width=0.23\linewidth]{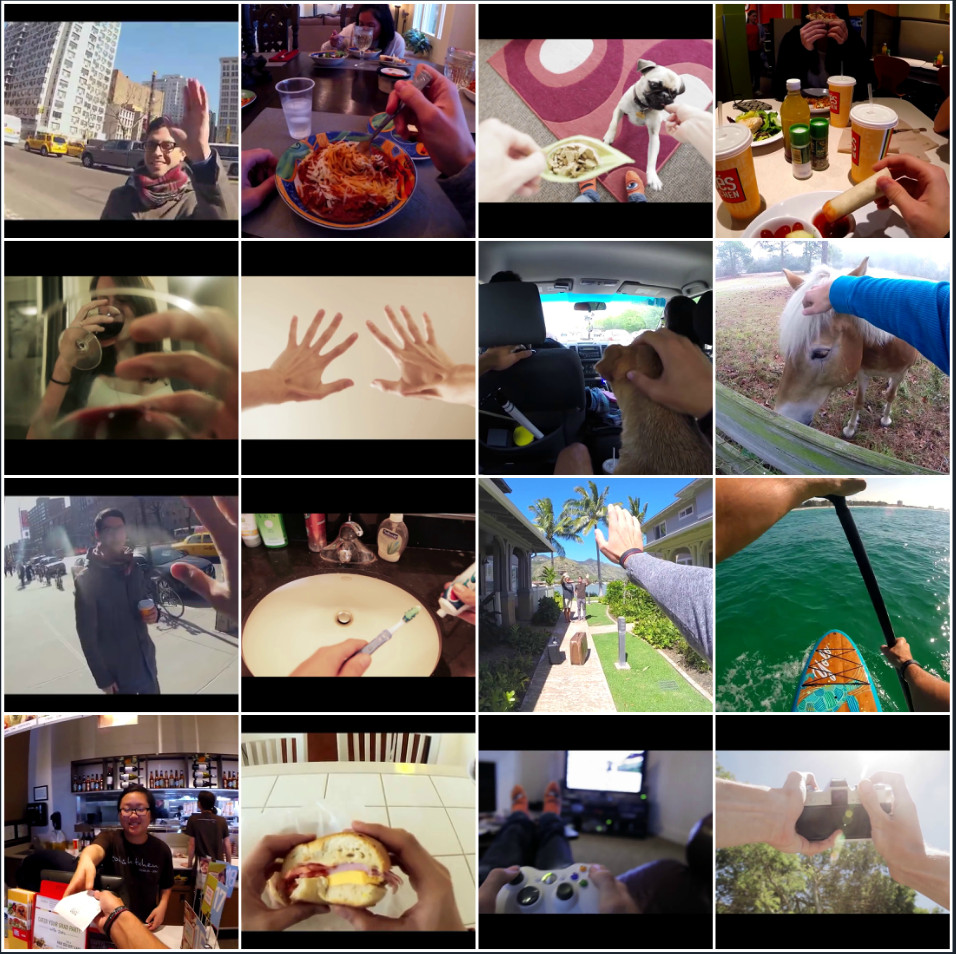}} &
\subcaptionbox{GTEA}{\includegraphics[width=0.23\linewidth]{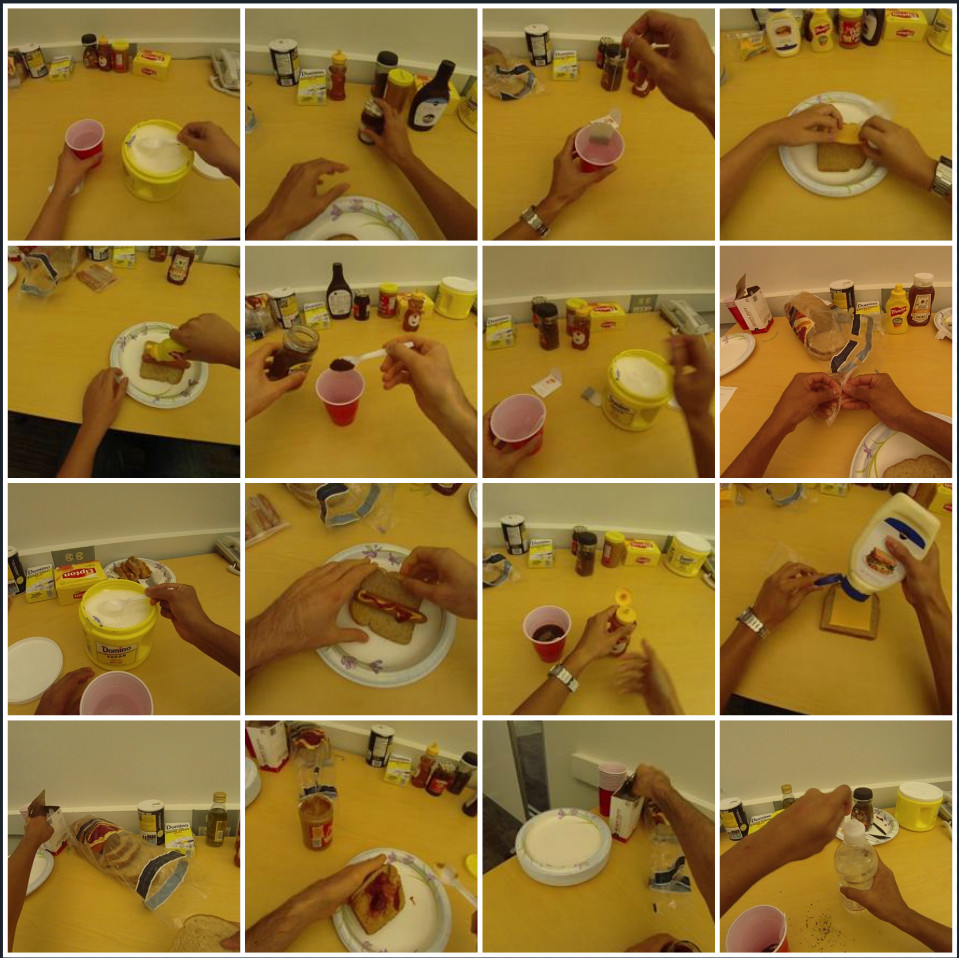}} &
\subcaptionbox{HandOverFace}{\includegraphics[width=0.23\linewidth]{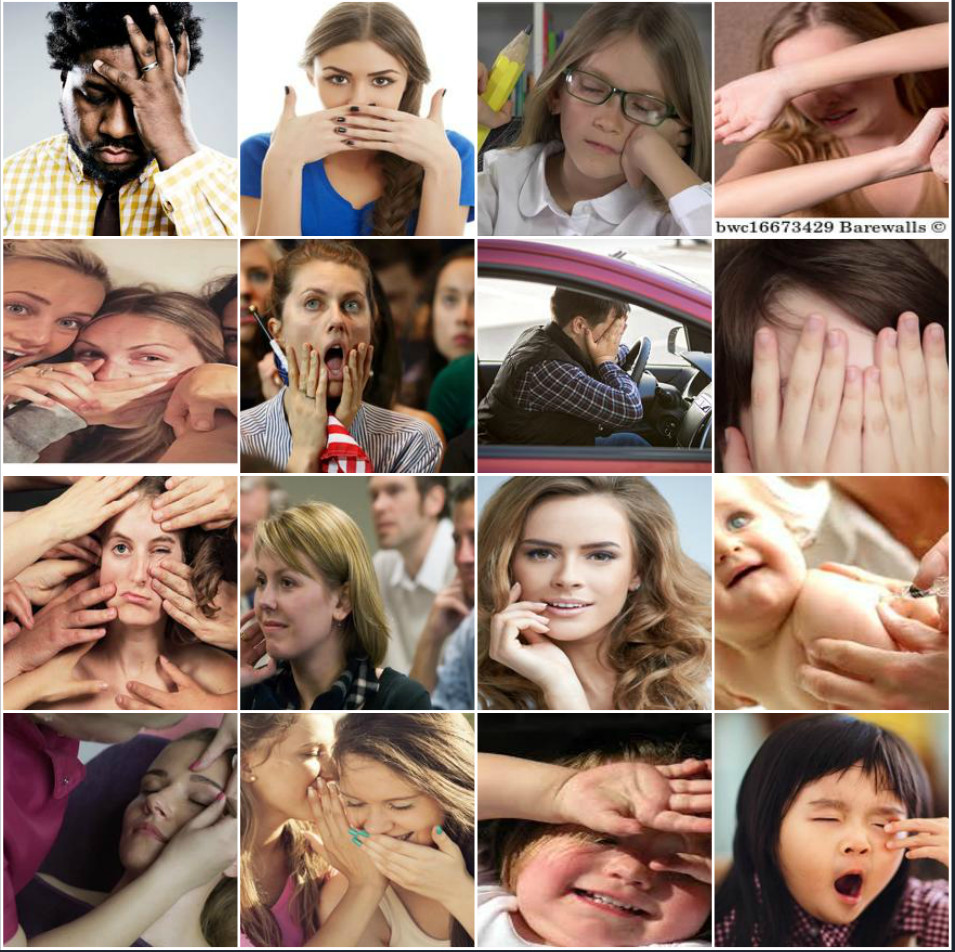}}
\end{tabular}
\vspace{-10pt}
 \caption{\small Sample images from 4 hand segmentation datasets including EgoHands, EYTH, GTEA and HOF, used in this paper.  }
\label{fig:datasets-samples}
\vspace{-5pt}
\end{figure*}

\begin{figure*}[t]
\centering
\includegraphics[width=1\linewidth]{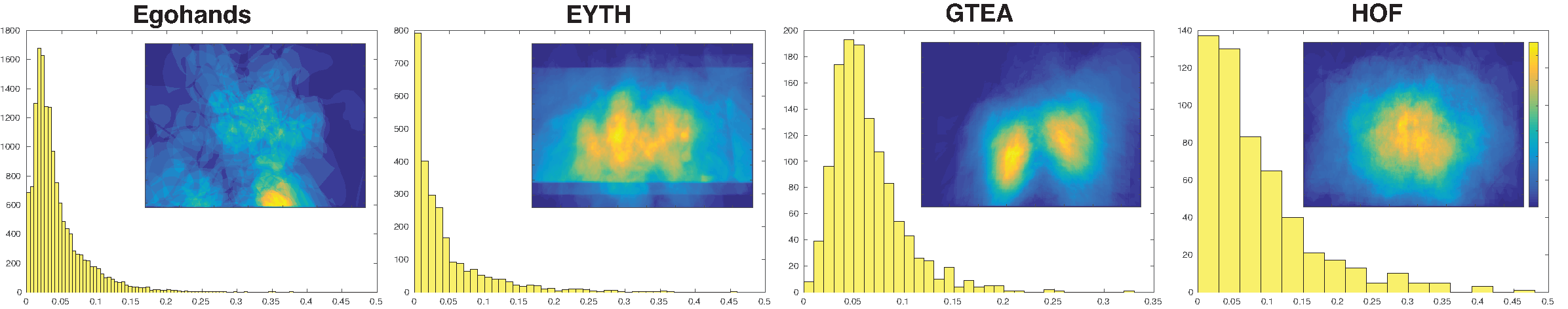}
\vspace*{-10pt}
 \caption{\small Visualizing the heatmaps for spatial hand occurrence in each dataset along with the histogram for bounding box area of hand relative to the image size. We can see that EYTH has most variation in hand location as compared to EgoHands (hands mostly occur at the bottom), GTEA (hands mostly occur in the same region) and HOF (hands mostly appear towards the center). After EYTH, HOF has more variation in hand location as well as in hand size. Histograms tell about the distribution of hand size (in terms of area) in these datasets.}
\label{fig:datasets}
\vspace{-10pt}
\end{figure*}

Several works have addressed egocentric hand segmentation. Ren and Gu~\cite{ren2007tracking} and Fathi et al.~\cite{fathi2011learning} proposed a method to find  regions with irregular optical flow patterns that may correspond to hands. The assumption here is that the background is static which is quite true in realistic daily egocentric videos when the person interacts with other objects or people. Li and Kitani~\cite{li2013pixel} proposed an illumination-aware approach that chooses the best local color feature for each environment using scene-level feature probes. However, they assumed that there are no social interactions in videos, so that all hands in the video belong to the egocentric viewer. Lee et al. \cite{lee2014hand} proposed an approach to detect hands in social interactions in egocentric videos. They also proposed a probabilistic graphical model to utilize spatial arrangements to disambiguate hand types (i.e., self vs. other). However, they only considered interactions in laboratory style conditions.

Most relevant to our work is Bambach et al.~\cite{egohands2015iccv}. They proposed a skin-based approach that first generates a set of bounding boxes that may contain hand regions and then use CNNs to detect hands, finally using GrabCut to segment them. They also attempted to determine hand types as well as predicting the activities from hand regions. Further, 
they introduced a new large dataset, EgoHands, including 48 first-person videos of people interacting in realistic environments, with pixel-level ground truth for over 15,000 hand instances. We use their method as our baseline. Recently, some large-scale RGB-D hand datasets have also been introduced to study hand pose tracking and estimation in depth images (e.g.,~\cite{yuan2017bighand2,garcia2017first}).


\noindent \textbf{Third-person hand segmentation.}
Some works have addressed hand detection in videos recorded from third-person or surveillance cameras. For example, Mittal et al.~\cite{mittal2011hand} used deformable part models and skin heuristics to detect hands. Few other recent works have investigated 3D hand pose estimation and hand-object interaction using hand segmentation~\cite{zimmermann2017learning, kang2017hand}.~\cite{cote2006comparative, zabulis2009vision} present comparative studies of hand segmentation methods from the pre deep learning era. 
\section{Analysis Plan} \label{analysis}
\vspace*{-5pt}
In this section, we first describe details of the datasets 
used in our analysis, followed by our approach for hand segmentation 
and hand-based activity recognition. 


\subsection{Hand Datasets} \label{hand_datasets}
\vspace*{-5pt}
We employ five datasets in our analysis. Two of them are already available (EgoHands and GTEA), and we contribute 3 additional datasets including EgoYouTubeHands, HandOverFace and EgoHands+. The first four will be used for the segmentation task and the last one for activity recognition. We used LabelMe \cite{russell2008labelme} toolbox to annotate hands in our datasets. For hand segmentation, we follow \cite{egohands2015iccv} and annotate hands only till wrist. For activity recognition, we labeled 8 videos from EgoHands dataset for activities at hand-pose level. Fig.~\ref{fig:datasets-samples} shows sample frames from these datasets.
\vspace*{-15pt}
\subsubsection{EgoHands dataset} 
\noindent This dataset~\cite{egohands2015iccv} is unique 
because it captures interactions among actors. It has 48 videos recorded with a Google glass. Each video has two actors doing one of the 4 activities: \emph{playing puzzle, cards, jenga or chess}. These videos are recorded in 3 different environments: \emph{office, courtyard and living room}. The EgoHands dataset has pixel-level ground truth for over 15000 hand instances. Each video has 100 manually annotated frames, 4800 frames ground truth in total. The original work randomly splits these videos into training, validation and test sets with the ratio of 75\%, 8\% and 17\%, respectively. We used the same split in our work. Fig.~\ref{fig:datasets} shows the heat map for spatial occurrence of hands in this dataset. As can be seen, most of the time hands occur at the bottom of the egocentric videos.
\subsubsection{EgoYouTubeHands (EYTH) dataset}
\noindent One limitation of existing datasets is that they are collected in constrained environments. Ideally, we would like  
to detect any hand in first-person videos recorded in unconstrained daily settings. To meet this objective, we need pixel-level hand annotations in real world images and/or videos. Thus, we downloaded three egocentric videos (3-6 minute long) from YouTube in which users are doing different activities and are interacting with others. 
We annotated every 5th frame in these videos. 
Our dataset has $\sim$1290 frames with pixel-level hand annotations, with variation in environment, number of participants, hand size, etc. Fig.~\ref{fig:datasets-samples} shows some images. Hands visible through transparent objects such as glass are annotated as per VOC 2009 annotation guidelines as complete object considering reflections as occlusion. This dataset has 2600 hand instances, with approx. 1800 first-person hand instances and approx. 800 third-person hands. The heatmap for EYTH in Fig. \ref{fig:datasets} shows that this dataset has most variations in hand locations. Also the histogram of this dataset has most images for smallest hand size. See Table \ref{tab:stats2} for detailed breakdown of statistics.
\vspace*{-5pt}
\subsubsection{GTEA dataset}
\vspace*{-5pt}
\noindent  GeorgiaTech Egocentric Activity dataset (GTEA) by \cite{delving} has 7 daily activities performed by 4 subjects. Videos are collected in the same environment for the purpose of activity recognition. 
 It does not capture social interactions and is collected under static illumination conditions annotated at 15 fps for 61 action classes. We use this dataset for hand segmentation. The original dataset has 663 images with pixel-level hand annotations considering hand till arm. Following \cite{cai2016understanding}, we also cropped out the arm part for our work.

\vspace*{-5pt}
\subsubsection{HandOverFace (HOF) dataset}
\vspace*{-5pt}
We collected 300 images from the Web in which faces are occluded by hands to study how skin similarity can affect hand segmentation. Some examples are shown in Fig.~\ref{fig:datasets}. This dataset has pixel-level annotations for hands along with the hand type: \emph{left or right}. HOF has images for people from different ethnicities, age, and gender. See Table \ref{tab:stats2} for statistics. HandOverFace dataset has a fair share of both big and small hands. Also, visualization of hand maps across this dataset tells us that hands are mostly appearing towards the center of images (See Fig~\ref{fig:datasets}).
\vspace*{-15pt}
\subsubsection{EgoHands+ dataset}
\vspace*{-5pt}
\noindent Humans are good at guessing the action being performed by just looking at the hand pose. The original EgoHands dataset is labeled at frame-level for 4 activities. To investigate the role of hand masks in hand-pose based activity recognition, we needed action annotations at a finer level. Thus, we annotated a small subset of 8 videos (800 frames), 2 from each coarse-level activity for outdoors (courtyard) in the EgoHands dataset at hand-pose level. We labeled each hand pose with one of the following 16 activities: \textit{holding, picking, placing, resting, moving, replacing, thinking, pulling, pushing, stacking, adjusting, matching, pressing, highfive, pointing, and catching}. Ambiguous hand poses are annotated with multiple possible labels(e.g., picking and placing are sometimes difficult to be inferred at a single frame-level). Additionally, most of these actions are based on the object type. For instance, adjusting and matching are actions relevant to puzzle pieces, pulling and pushing are relevant to jenga blocks, and so on. However, actions like holding, thinking, resting and highfive are general actions.

\begin{figure}[t]
\begin{center}
\includegraphics[width=\linewidth]{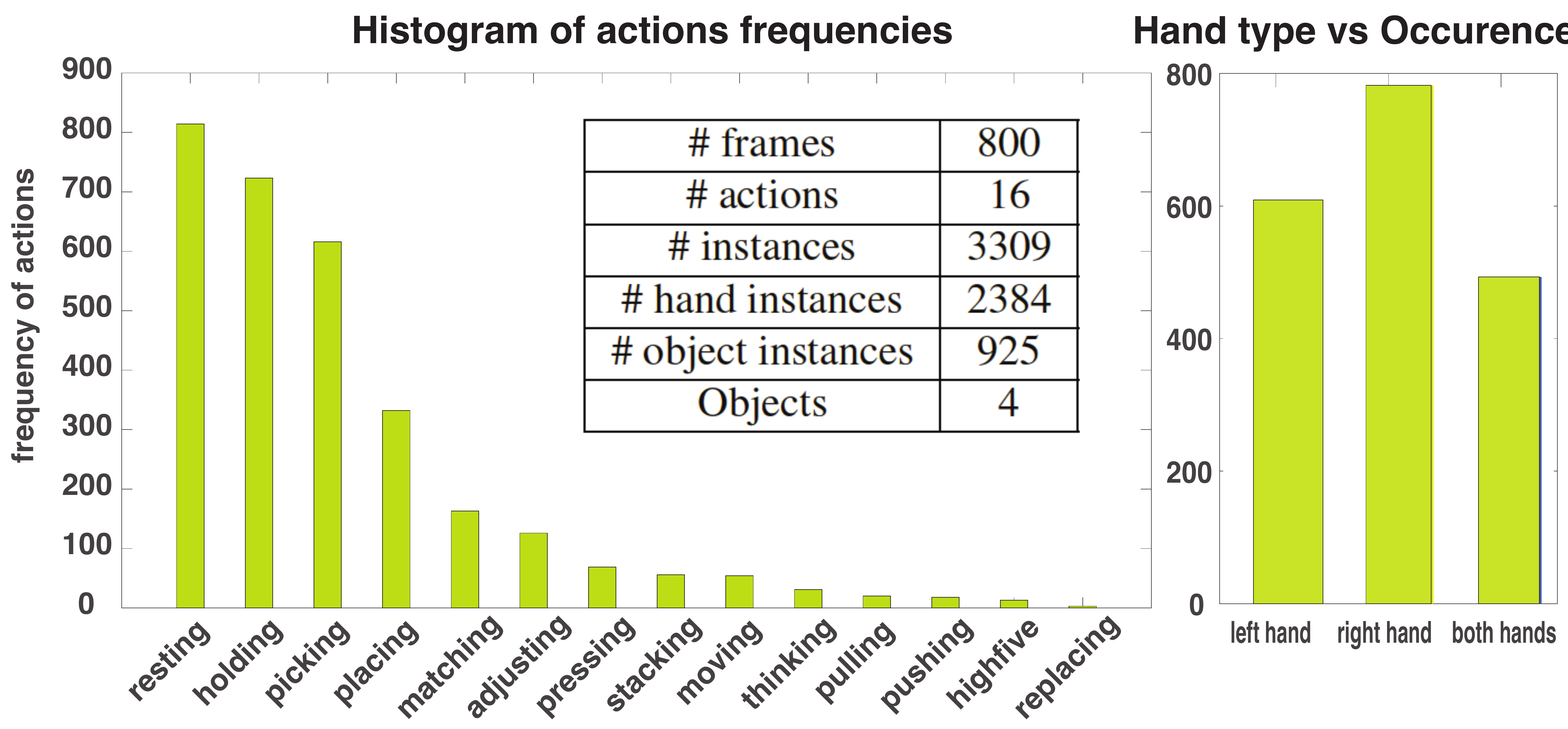}
\end{center}
\vspace{-10pt}
   \caption{\small Left) Histogram of activities occurrence in EgoHands+ dataset. Fine-level activities include holding, picking, placing, resting, moving, replacing, thinking, pulling, pushing, stacking, adjusting, matching, pressing, high-five, pointing, and catching.
   Left: Inset) Some statistics for EgoHands+ dataset. Objects include cards, chess\_piece, jenga\_block and puzzle\_piece. We used this dataset for fine-level activity recognition.
   Right) Histogram of hand type occurrence in EgoHands+ dataset
   }
\label{fig:ego1}
\vspace{-15pt}
\end{figure}

We call these additional annotations as the EgoHands+ dataset. Fig.~\ref{fig:ego1} shows occurrence of each action in the EgoHands+ dataset. Actions like holding, picking, placing and resting are few of the most occurring ones in this dataset. Fig. \ref{fig:ego1} also shows the break down of all actions based on their hand-type i.e., whether the participant used her \textit{left, right, or both hands}. Fig. \ref{fig:ego1} also shows statistics over the EgoHands+ dataset.

\begin{table}[t]
\footnotesize
\renewcommand{\arraystretch}{.6}
  \centering \setlength{\tabcolsep}{.56\tabcolsep}   
    \begin{tabular}{lccccc}
       \toprule
                \multicolumn{1}{c}{\textbf{Stat}} & \multicolumn{4}{c}{\textbf{Dataset}}  \\
                \cmidrule(lr){2-5}

          & EgoHands  & EgoYouTube.  & GTEA  & HandOverFace    \\
          
        \midrule
    frames  & 4800 & 1290 & 663 & 300 \\
    \midrule
    hand instances  & 15053 & 2600 & 1231 & 507  \\
    \midrule
    1st-P instances & 5976 & 1811 & 1231 & \textbf{--}  \\
    3rd-P instances  & 9077 & 789 & \textbf{--} & 507  \\
    \midrule
    1st-P left & 2560 & 920 &600 &  \textbf{--}\\
 	1st-P right  & 3416 & 631 & 192& \textbf{--}\\
	\midrule
 	3rd-P left  & 4567 & 255 &  \textbf{--} & 230 \\
 	3rd-P right  & 4510 & 534 &  \textbf{--} & 277 \\
    \hline
    \end{tabular}
    \vspace{-5pt}
    \caption{\small Statistics of hand datasets used in this work. 1st-P and 3rd-P stand for first-person and third-person, respectively.}
    \label{tab:stats2}
    \vspace{-20pt}
\end{table}

\subsection{Hand Segmentation} \label{hand_segmentation}
\vspace{-5pt}
We treat hand segmentation as a semantic segmentation task, in contrast to Bambach \textit{et al.} who formulated it as object detection. Semantic segmentation assigns one label, from a well defined set of class labels, to each pixel~\cite{oliveira2016deep}. Similarly, we interpret the hand detection problem as a dense prediction problem where we want to detect every pixel that belongs to a hand (i.e., binary segmentation).

\noindent \textbf{Adopting RefineNet for hand segmentation:}
Fully convolutional networks~\cite{long2015fully} have been successfully used for dense prediction tasks. 
In recent years, deep residual nets have been used as the backbone in several models such as PSPNet \cite{zhao2016pyramid} and RefineNet \cite{lin2016refinenet}. RefineNet is a multi-path refinement network which exploits all the features at multiple levels along the down sampling path. A RefineNet block typically consists of Residual Convolution Units (RCU), Multi-Resolution Fusion of features coming from the RCU blocks and Chained Residual Pooling. RefineNet is a cascaded architecture of multiple RefineNet blocks which is based on Residual net features. It computes features from ResNet at different levels and fuses them to produce a high resolution prediction map. We picked off-the-shelf 4-cascaded RefineNet model to evaluate it on the EgoHands dataset. We performed off-the-shelf evaluation of leading semantic segmentation methods (\cite{chen2016deeplab} and \cite{zhao2016pyramid}) on the EgoHands dataset, and find that RefineNet gives better results than other models. Since \cite{chen2016deeplab} and \cite{zhao2016pyramid} were trained on PASCAL VOC for person class, we evaluated their performance only on hand regions to give them an advantage, but off-the-shelf RefineNet beats \cite{chen2016deeplab} with a significant margin ($\sim$17\%) and \cite{zhao2016pyramid} with a slight difference($\sim$2\%). Therefore, we chose it for our analysis on hand segmentation datasets. 

We used RefineNet-Res101 pre-trained on Pascal Person-Part dataset in all our experiments. We used pre-trained RefineNet-Res101 with a new classification layer with 2 classes: \textit{hand} and \textit{no hand}. We trained the model on EgoHands, EYTH, GTEA, and HOF datasets. RefineNet-Res101 uses feature maps from ResNet101. After fine-tuning, we perform multi-scale evaluation for scales: [0.6, 0.8, 1.0] as mentioned in \cite{lin2016refinenet} which gives us consistently better results than single scale evaluation. On EgoHands dataset, RefineNet significantly outperformed the baseline (See Fig. \ref{fig:long}). Thus, we used this fine-tuned model as our pixel-level hand detector. 

\noindent \textbf{Cross dataset evaluation.}
We believe that along with a robust segmentation method, appropriate hand segmentation datasets are also important for accurately segmenting hands in the wild.
To find the essential ingredients for a robust hand segmentation setup, we measure generalization capabilities of RefineNet across datasets. After training RefineNet on each dataset, we test it across datasets to study its generalization capability (See Table \ref{tab:diff-datasets-results}). 

\noindent \textbf{Further refinement with CRFs.} CRFs are well known for being useful in refining pixel-level predictions for computer vision problems such as saliency detection and semantic segmentation. Our initial task is to segment hands from an input image which is a binary semantic segmentation problem and thus, is quite similar to foreground/background estimation. We employed unary and pairwise potentials from \cite{mehrani2010saliency} to refine initial maps obtained by our hand detector (Fig. \ref{fig:long} shows results on EgoHands dataset and Fig. \ref{fig:crf_results} shows results on all datasets). 

\noindent \textbf{Small hands vs. big hands.} For further analysis, we selected images with small hands and big hands from all datasets based on a threshold on bounding box area of hand relative to the image, and evaluated each trained model on small and big hands from the same dataset, respectively.

\subsection{Hand-based Activity and Action Recognition.} \label{activity_recognition}
\vspace{-5pt}
Hand regions can tell us about the activities a person is doing. Bambach \textit{et al.} showed a correlation between hand regions and activity being performed. Here, we extend their task. Given a single hand map, we aim to predict the fine-level action (1 out of 8 over EgoHands+ dataset). 
We consider activity recognition at two levels: \textbf{coarse-level} - where activity label is available at the frame level, and \textbf{fine-level} - where we have action labels for each hand region.

\noindent \textbf{Coarse-level activity recognition.} Similar to \cite{egohands2015iccv}, we also perform hand-based activity recognition but we aim to test it using better segmentation maps that we generate to see if better segmentation helps activity recognition. 
Note that EgoHands dataset has frame-level annotations for different activities. The task is to classify activity only using hands without any background information. For coarse-level activity recognition, we used the same classification model used by \cite{egohands2015iccv} and trained it on EgoHands dataset to reproduce their results. We find that better hand maps lead to better activity recognition. %

\noindent \textbf{Fine-level action recognition.} We extended Bambach \textit{et al.}'s work to finer level as different hand poses are used for different actions like holding or writing. Given a single hand instance, we aim to tell what fine-level action is being performed. For instance, while playing cards is a coarse activity, a person can use his hands for \textit{picking} and \textit{placing} cards which are fine-level actions. Since, few actions like \textit{highfive, catching, replacing, and pushing} rarely occur, we trained a CNN for 8 most frequent action classes which uses a single hand instance to classify which fine-level action is being performed. \noindent We trained the same CNN~\cite{NIPS2012_4824} in all of our experiments related to activity/action recognition, except that the last layer changes according to the number of classes. 

\begin{table*}[t]
\small
  \centering
  \renewcommand{\arraystretch}{.9}
  
    \begin{tabular*}{\textwidth}{l @{\extracolsep{\fill}} ccccccccccccccc}
        \toprule
                \multicolumn{1}{c}{\textbf{Dataset}} & \multicolumn{3}{c}{\textbf{EgoHands}} & \multicolumn{3}{c}{\textbf{EYTH}} & \multicolumn{3}{c}{\textbf{GTEA}}  & \multicolumn{3}{c}{\textbf{HOF}}   \\
                \cmidrule(lr){2-4}
\cmidrule(lr){5-7} \cmidrule(lr){8-10} \cmidrule(lr){11-13}
          & mIOU  & mRec  & mPrec  & mIOU  & mRec  & mPrec  & mIOU  & mRec  & mPrec & mIOU & mRec & mPrec\\         
        \midrule
    EgoHands  & \textbf{0.814}&\textbf{0.919}&\textbf{0.879} & 0.428&0.615&0.550 & \textcolor{blue}{0.774}&\textcolor{blue}{0.834}&\textcolor{blue}{0.904} & 0.503&0.738&0.619 \\
    EYTH & \textcolor{blue}{0.670} & \textcolor{blue}{0.768}  & \textcolor{blue}{0.841} & \textbf{0.688}&\textbf{0.776}&\textbf{0.853} & 0.666&0.700&0.920 & \textcolor{blue}{0.528 }&\textcolor{blue}{0.653}&\textcolor{blue}{0.722} \\
    GTEA & 0.152&0.307&0.204 & \textcolor{blue}{0.440}&\textcolor{blue}{0.569}&\textcolor{blue}{0.614} & \textbf{0.821}&\textbf{0.869}&\textbf{0.928} &  0.263  & 0.880 & 0.276 \\
    HOF & 0.578&0.701&0.780 & 0.423&0.497&0.667 & 0.431&0.450&0.879 & \textbf{0.766}&\textbf{0.882}&\textbf{0.859 }\\   
        \bottomrule
    \end{tabular*}
    \vspace*{-5pt}
    \caption{\small Hand segmentation results on 4 datasets using RefineNet trained on one dataset and tested on others. Numbers in bold text show the best results on a particular dataset, whereas the blue font shows second best results for that dataset (Vertical: Train; Horizontal: Test).}
    \label{tab:diff-datasets-results}
        \vspace{-10pt}
\end{table*}

 \begin{figure}[t]
\begin{center}
\footnotesize
\renewcommand{\arraystretch}{.9}
\includegraphics[width=1\linewidth]{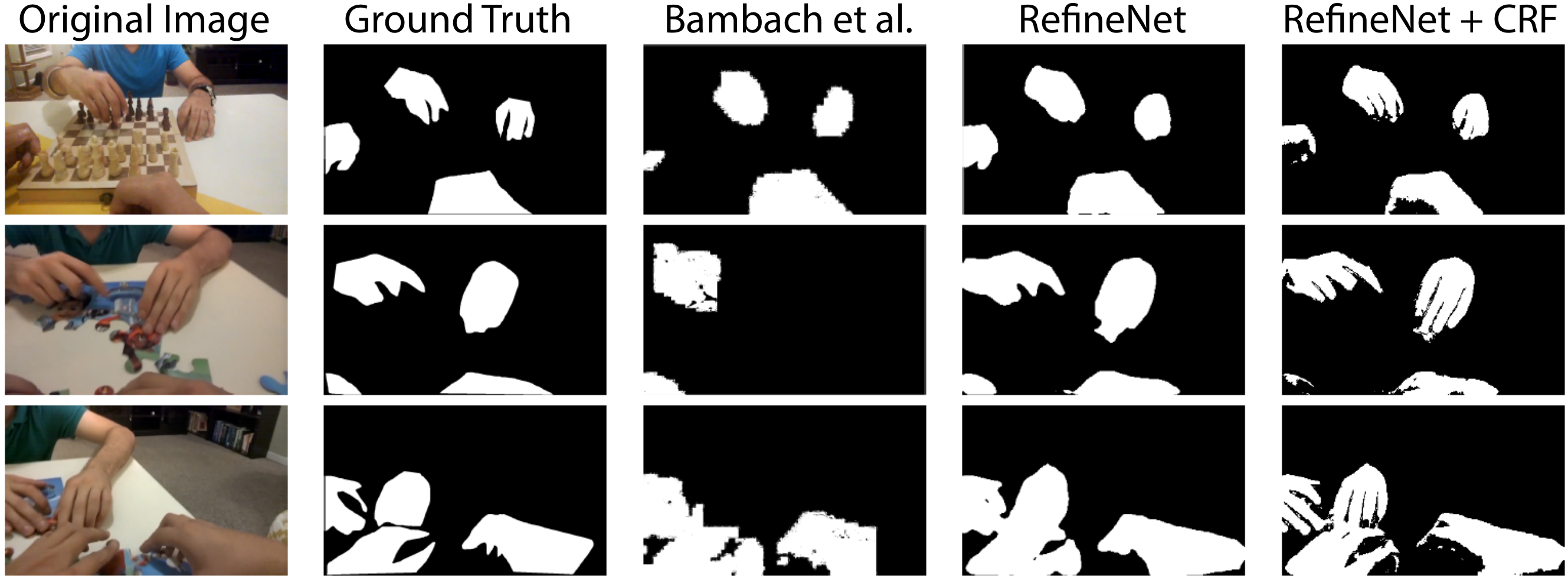}
\end{center}
\vspace*{-15pt}
   \caption{\small Few examples of qualitative results for hands segmentation on EgoHands dataset using the baseline \cite{egohands2015iccv}, RefineNet, and RefineNet+CRF methods. }
\label{fig:long}
\label{fig:crf}
\vspace*{-15pt}
\end{figure}
	
\vspace*{-5pt} 
\section{Experiments and Results} \label{results}


\noindent \textbf{Evaluation metrics.} For \textit{hand segmentation}, we report three metrics: pixel-level mean Intersection over Union (mIOU), mean Precision, and mean Recall. For \textit{activity recognition}, we report classification accuracy.



\subsection{Segmentation Results} \label{seg_results}
\noindent Our baseline method~\cite{egohands2015iccv} detects hand instances (bounding boxes), aggregates them, and then runs GrabCut to generate a segmentation
map. This map can be converted to binary and compared with our model. Whereas, we use RefineNet-Res101 pre-trained on Pascal Person-Part dataset. We chose this model since it already has been trained to parse human hands on Person-Part dataset. The model parses human hand till elbow whereas we consider hands till wrist. For training RefineNet, we used a learning rate of 5e-5 with an increased learning rate by a factor of 10 for our binary classification layer. All models were trained till convergence, and multi-scale evaluation is used with three scales: [0.6, 0.8 and 1.0].

In the following, we first report results over the egocentric video datasets, and then proceed to some analyses.


\noindent \textbf{1) EgoHands dataset:} 
For fine-tuning pre-trained RefineNet-Res101 on EgoHands dataset, we used the same data split as in \cite{egohands2015iccv} with 3600, 400 and 800 images for training, validation and testing, respectively. We fine-tuned RefineNet-Res101 on EgoHands dataset for 80 epochs. After multi-scale evaluation, we obtained an mIOU of 81.4\% outperforming the baseline score i.e., 55.6\%. We also achieved pixel-level mPrecision of 87.9\% and mRecall of 91.9\%. Fig. \ref{fig:crf} shows qualitative results on EgoHands.

\noindent \textbf{2) EgoYouTubeHands (EYTH) dataset:}
We randomly split the images into 60\%-20\%-20\% split with 774 training images, 258 validation and 258 testing images, respectively. We trained RefineNet on EgoYouTubeHands dataset for 85 epochs and achieved 68.8\% mIOU, 85.3\% mPrecision and 77.6\% mRecall. Among all datasets, this is the lowest mIOU we obtained due to the challenging nature of the EYTH dataset. 

\noindent \textbf{3) GTEA dataset:} Similar to \cite{fathi2011understanding} and \cite{fathi2013modeling}, we also used images from subjects 1, 3 and 4 for training, and subject 2 for testing. Training images were further split into 80\%-20\% with 367 training images and 92 validation images, whereas the test set has 204 images. RefineNet was trained on GTEA for 92 epochs till its convergence which resulted in 82.1\% mIOU on the test set. RefineNet obtains highest mean precision of 92.8\% on GTEA and mRecall of 86.9\%.

\begin{figure}[t]
\begin{center}
\includegraphics[width = 1\linewidth,height=2.5in]{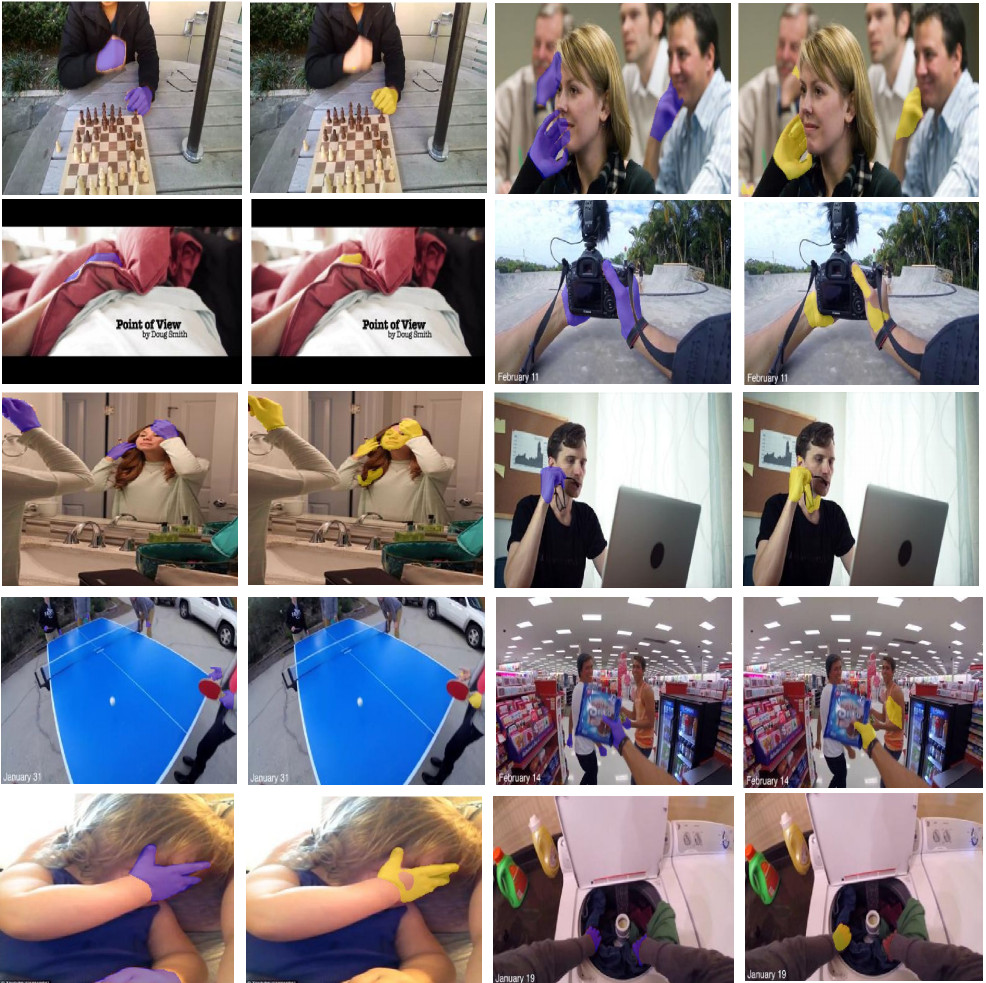}
\end{center}
\vspace{-15pt}
\caption{\small \textbf{Failure cases.} Some challenging cases (from all datasets) where segmentation model fails (mIOU $<$ 0.6). Images with blue masks show ground truth and yellow masks show predictions. Row 1 shows failure cases for motion blur, row 2 indicates failure cases due to occlusion, row 3 shows examples when skin appearance occlusion happens, row 4 shows cases when model fails for small hands, and the last row shows examples where lighting situation was either too bright or insufficient.}
\label{fig:failure_cases}
\vspace{-18pt}
\end{figure}

\noindent \textbf{Performance analysis on HandOverFace (HOF).}
The HandOverFace is a small dataset of 300 images for hand over face occlusion having pixel-level hand annotations along with labels for hand type i.e., \textit{left} and \textit{right} (collected from publicly available images from Web). We randomly split this dataset into 80\%-20\% train-test split. Training images were further split into 200 training and 40 validation images. We used 60 images for testing. The model was trained for 61 epochs and gives 76.6\% mIOU, 85.9\% mPrecision and 88.2\% mRecall.

\noindent \textbf{Cross dataset generalization.} 
To measure cross dataset performance, we applied RefineNet trained on each dataset to other datasets. We find that RefineNet, trained on EgoYouTubeHands dataset, generalizes the best across other datasets mainly due to its variations in lighting conditions, small and big hand instances, occlusions, people appearing in the field of view in unconstrained manners, etc. Interestingly, RefineNet trained on EYTH gives second best results on EgoHands and HOF datasets. EgoHands dataset has the setting where actors interact with each other and HOF dataset has images with similar appearance occlusions, both of which happen quite often in EYTH. 


GTEA, although gives second best results on EYTH. The generalization power of RefineNet when trained on GTEA is poor among all other datasets. This is because GTEA does not have images for social interactions and was collected in a static environment. Hence, it is difficult for RefineNet trained on GTEA to perform well in the wild.

Surprisingly, RefineNet when trained on HOF, despite being the smallest dataset, still generalizes better than GTEA. This suggests that if a network sees similar appearance occlusions repeatedly, it will learn to distinguish hands from other body parts.
Table \ref{tab:diff-datasets-results} summarizes these results.

\noindent We also studied examples where our trained models underperform (below 0.6 mIOU), and report possible causes along with failure cases in Fig. \ref{fig:failure_cases}.

\begin{figure}[t]
\centering
\includegraphics[width=1\linewidth]{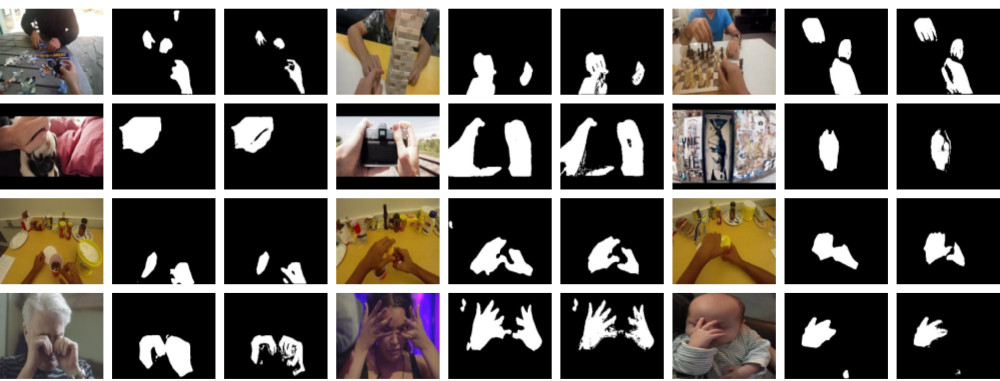}
\vspace*{-15pt}
 \caption{\small \textbf{CRF refinement results}. Column 1 shows input image, column 2 shows classification results, and column 3 shows results after CRF. The same order is followed for rest of the columns. Rows 1, 2, 3 and 4 show results for EgoHands, EYTH, GTEA and HOF, respectively. CRF sometimes helps remove false positives (see row 1, example 1 and row3, example 2). It also improves hand maps when both hands are intersecting (see row 3, example 2 and 3). Overall, CRF refines hand details (fingers, pose, etc.) giving more visually appealing results.}
\label{fig:crf_results}
\vspace{-10pt}
\end{figure}

\noindent \textbf{Performance analysis of CRF.} After computing unary and pairwise potentials from test images and classifier results, we used GCMex~\cite{Fulkerson2009,Boykov2001,Boykov2004,Kolmogorov2004} for energy minimization. We find that, although CRF helps to visually improve fine-level details of hands like fingers, overall it slightly hurts them in terms of mIOU. Fig. \ref{fig:crf} shows qualitative results for some successful cases of applying CRF on EgoHands dataset. Among all datasets, CRF worked best for GTEA where it negligibly affected the mIOU (a drop of 0.5\%) giving us visually appealing results. Particularly in case of hand-to-hand occlusion, when the classifier failed to detect them as separate hands, CRF successfully refined them as multiple hand instances. Fig. \ref{fig:crf_results} shows qualitative results for all datasets after using CRF.

\begin{figure}[t]
\centering

\includegraphics[width=1\linewidth]{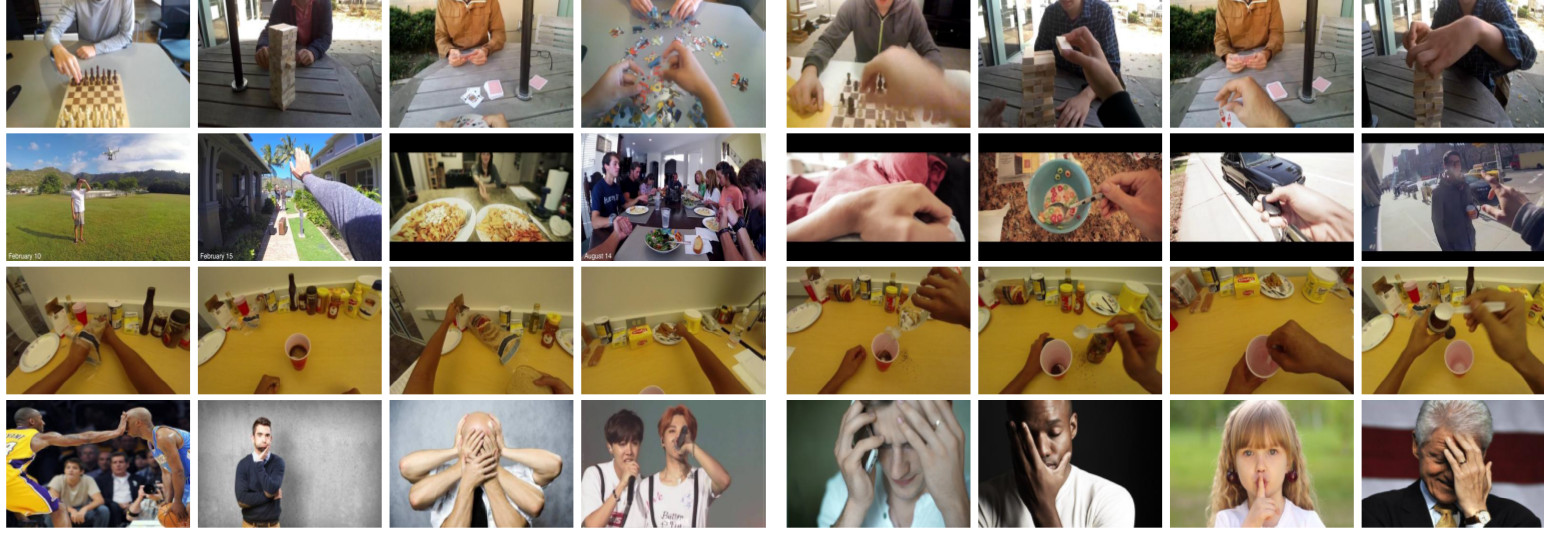}

\vspace*{-5pt}
 \caption{\small Examples of small hands (left) and big hands (right). Rows 1, 2, 3 and 4 show examples from EgoHands, EYTH, GTEA and HOF datasets, respectively.}
\label{fig:small_vs_big}
\vspace{-20pt}
\end{figure}

\noindent \textbf{Performance analysis on small vs. big hands.} Apart from hand segmentation, we were also interested in investigating the relationship between hand size and segmentation accuracy. We selected images with small hands and big hands from test sets for all four datasets (See Fig. \ref{fig:small_vs_big} for few examples), and evaluated their respective models on those test images.  
We consider hands with relative area less than 0.015 as small hands, except for GTEA where we thresholded hands smaller than 0.025 as small hands. The reason for choosing higher threshold for GTEA is because this dataset has very few images with smaller hands. Please see Fig. \ref{fig:datasets} for histogram of relative area of hands. Hands with relative area larger than 0.15 are treated as big hands for all datasets. For evaluating these images from each dataset, we used the model trained on the same dataset. For instance, for EgoHands dataset, we used model trained on EgoHands dataset, and so on.

Our results show that the mIOU is consistently lower for small hands on all datasets, except for EgoHands, where the mIOU for small hands is slightly higher than big hands. The possible reason for this is that, in EgoHands dataset, most hands are smaller in size. In addition to that, EgoHands has a constrained setting of first-person and third-person interactions, due to which, same image with small hand for third-person may have a big hand for first-person (See Fig \ref{fig:datasets}). Therefore, it makes the performance better than big hands for the chosen thresholds, but with a low margin(3.7\%). Whereas, for other three datasets, the mIOU for big hands is higher with a large margin($\sim$8\%-33\%), validating that small hands are more challenging for segmentation. Overall, we find that the model struggles with segmenting small hands as compared to big hands(See Table \ref{tab:small_big}).

\begin{table}[t]
\small
\renewcommand{\arraystretch}{.9}
  \centering \setlength{\tabcolsep}{.56\tabcolsep}   
    \begin{tabular}{lcccccc}
        \toprule
                \multicolumn{1}{c}{\textbf{Datasets}} & \multicolumn{3}{c}{\textbf{Small Hands}} & \multicolumn{3}{c}{\textbf{Big Hands}}   \\
                \cmidrule(lr){2-4}
\cmidrule(lr){5-7} 
          & mIOU  & mRecall  & mPrec  & mIOU  & mRecall  & mPrec   \\
          
        \midrule
    EgoHands  & \textbf{0.787} & \textbf{0.917} & \textbf{0.850} & 0.750 & 0.925 & 0.802  \\
    EYTH  & 0.537 & 0.643 & 0.693 & \textbf{0.867} & \textbf{0.914} & \textbf{0.944} \\
    GTEA & 0.732 & 0.787 & 0.913 & \textbf{0.894} & \textbf{0.927} & \textbf{0.962} \\
    HOF & 0.713 & 0.866 & 0.808 & \textbf{0.792} & \textbf{0.932} & \textbf{0.840} \\
    \midrule
 
    \end{tabular}
    \vspace*{-10pt}
    \caption{\small Performance analysis on small vs. big hands.}
    \label{tab:small_big}
    \vspace{-17pt}
\end{table}

\vspace{-5pt}
\subsection{Activity and Action Recognition Results} 
\vspace{-5pt}
\noindent \textbf{Coarse-level activity recognition:} Our baseline~\cite{egohands2015iccv} trained a Caffe \cite{jia2014caffe} based CNN as a 4-way classifier (coarse level) for 4 activities: \textit{cards, chess, jenga and puzzle}. To reproduce their results, we trained the same Caffe based CNN~\cite{AlexNet} on ground truth segmentation maps for 6K iterations. After training for 6K iterations, we tested the trained model on ground truth maps, prediction maps from baseline method and prediction maps produced by RefineNet trained on EgoHands. 

\begin{figure}[t]
\centering
\hspace{-17pt} \includegraphics[width=1.05\linewidth,height = 1.4in]{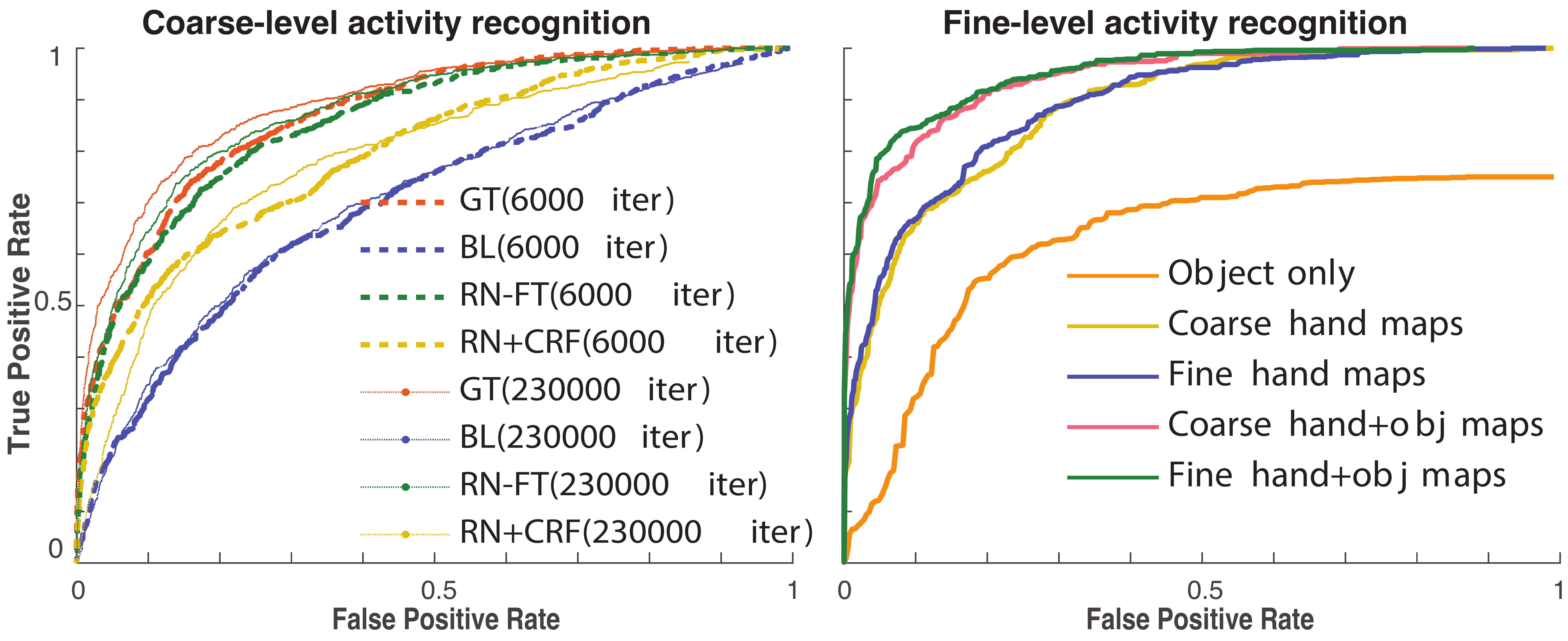}
\vspace*{-7pt}
 \caption{\small Average ROC curves for activity/action recognition. Left) Coarse level activity recognition for 4 actions: cards, chess, jenga and puzzle in EgoHands dataset. Right) Fine level action recognition for 8 most frequent actions in EgoHands+ dataset. We can see that detailed hand maps slightly improve accuracy over coarse hand maps for fine-level action recognition.}
\label{fig:roc}
\vspace{-15pt}
\end{figure}

Using the fine-tuned RefineNet, we obtain 13.6\% improvement over the baseline with recognition accuracy of 64.5\% which is closer to 66.5\% accuracy using ground truth hand maps. We also studied how well the action classifier performs on CRF- based maps and learned that it still performs better than the baseline by 5.3\%. See Table \ref{tab:activityrecognition} for detailed comparisons. We continued to train the network until it converged after 230K iterations and obtained an accuracy of 71.1\% using ground truth maps. While we were able to reproduce accuracy on ground truth maps as mentioned in \cite{egohands2015iccv}, testing it on baseline maps gives us lower accuracy than reported in the baseline paper. Therefore, we report the performance for their maps from their paper. Our converged model achieved 41\% accuracy on their hand maps. Fig. \ref{fig:roc} shows average ROC curves when we tested the model on ground truth and hand maps generated from baseline, RefineNet-FT and RefineNet+CRF after 6K and 230K iterations, respectively. We obtained the ROC curve for baseline from our trained model.  

\begin{figure}[t]
\centering
\includegraphics[width=0.48\textwidth]{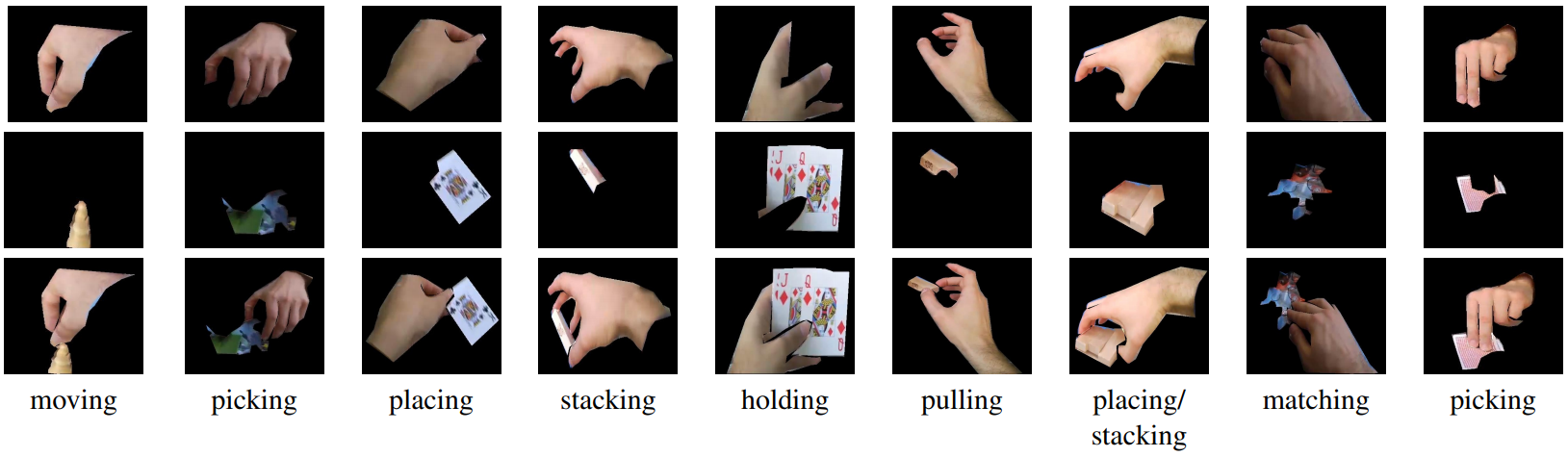}
\vspace{-20pt}
\caption{\small Sample images for few actions from EgoHands+ dataset focused on regions containing hand and/or object. Each column shows poses for different actions. Row 1 shows images used for training model on hand pose only, row 2 shows images for objects only, and row 3 shows images with hand+object.}
\label{fig:egoyou2}
\vspace{-15pt}
\end{figure}

\noindent \textbf{Fine-level action recognition:} To investigate action recognition at hand instance level, we additionally annotated a subset of EgoHands dataset for 16 hand-pose level actions and refer to it as EgoHands+ dataset. We then selected 8 most frequent actions (\textit{resting, holding, picking, placing, matching, pressing and stacking}) for training an 8-way CNN classifier with the same architecture used for coarse level activity recognition. We used base learning rate of 1e-4 with step size of 10K iterations. 
Additionally, we annotated same videos for finer hand maps and tried to segment hand details like fingers as much as possible. We then trained CNNs for 3 setups: 1) hands only, 2) object only, and 3) hand+object (See Fig.~\ref{fig:egoyou2}). We further split them on the basis of coarse hand maps and fine hand maps (where we have details about fingers).

We used the same split ratio as \cite{egohands2015iccv} for EgoHands+ dataset with the ratio of 75\%, 8\% and 17\% for training, validation and testing, respectively. Each input image was split for one hand map per image and then we trained a CNN using 5 setups: 1) Using coarse hand maps only, we achieved 58.6\% accuracy on test data where the chance level is 12.5\%, 2) Using fine-level hand maps slightly improves accuracy, 3\&4) Using objects along with both coarse and fine hand maps gives us the same results with accuracy of 77.3\%. As manipulated object along with hand improves accuracy by approx 18\%, we were curious to see how much the manipulated object carries information on its own. Thus, we trained a CNN on 5) Using objects only, we achieved an accuracy of 55.1\%. This suggests that objects carry useful information but not more than hands. Albeit, hands along with objects lead to a significant boost in performance. Fig. \ref{fig:roc} (right) shows ROC curves averaged over all actions for all 5 setups. 

To further explore the fine-level action recognition, we test the classifier trained on GT hand maps over predicted hand maps from RefineNet and achieve accuracy of 57.1\%. Testing on maps generated by RefineNet + CRF gives 54.4\% accuracy. We find that RefineNet maps give accuracy close to when using GT maps (59.2\%).

\begin{table}[t]
\small
  \centering \setlength{\tabcolsep}{.9\tabcolsep}   
    \begin{tabular}{lcc}
        \toprule

        \textbf{Method}  & \textbf{Acc. (6K iters)}  & \textbf{Acc. ($\sim$230K iters)}    \\          
        \midrule
    Ground truth    & 66.5\% & 71.1\%   \\
    
    Baseline hand maps~\cite{egohands2015iccv}   & 50.9\% & \textbf{--}  \\
    
    RefineNet-FT   & 64.5\%  & 68.4\%  \\
    RefineNet+CRF & 56.2\% &57.1\%   \\
   
	\bottomrule
    \end{tabular}
        \vspace{-10pt}
    \caption{\small Coarse level activity recognition on EgoHands dataset when network sees only hand maps. Activities include cards, chess, puzzle and jenga.}
    \label{tab:activityrecognition}
    \vspace{-20pt}
\end{table}


       
    
   

\vspace{-8pt}
\section{Discussion and Conclusion} \label{conclusion}
\vspace*{-8pt}

Accurate segmentation of hands is challenging yet useful for many applications, mainly in robotics and surveillance. We trained a hand segmentation model which gives improved results over the previous state of the art hand segmentation method~\cite{egohands2015iccv}. We also proposed 3 new datasets: 1) EYTH, a challenging dataset with real world settings, which is proved to be more versatile than existing egocentric datasets based on our results, 2) HandOverFace dataset, which is useful to study similar appearance occlusions when dealing with hands, and can help identify how we can deal with hand-to-skin occlusions, and 3) EgoHands+ dataset with action labels along with hand types (left, right, first-person, third-person) for each pixel-level annotated hand. For activity recognition, besides showing that improved hand maps improve recognition accuracy, we also reported baseline performance for fine-level action recognition and showed that even single hand instances are useful for finer level action recognition. Recent sophisticated deep networks are expected to give even higher performance.

Our work suggests some areas for improvement where even leading methods fail (e.g., hand-to-hand occlusion, small hands, poor lighting conditions, hand over face occlusions, etc). Along with models that handle these challenges, we also need large datasets with pixel-level annotations for hands in the wild. 
Conditional random fields although did not help us much quantitatively, but generated visually appealing segmentation maps. We experimented with DenseCRF on EgoHands dataset, but our preliminary results show no improvement. We will consider using some better high level information like superpixels to improve the results.

In summary, we took a deep look into the problem of hand segmentation in realistic unconstrained environments, proposed a model that outperforms the state of the art, introduced several datasets, and identified challenges that need to be addressed in future works. All code and data will be freely available to the public. 

\noindent \textbf{Acknowledgement.} We would like to thank NVIDIA for the donation of the Titan-X GPUs used in this work.


{\small
\bibliographystyle{ieee}
\bibliography{egbib}
}

\end{document}